\documentclass[sigconf]{acmart}

\usepackage{multirow}
\usepackage{amsmath} 
\renewcommand{\vec}[1]{\boldsymbol{#1}} 

\usepackage{graphicx}
\usepackage{epstopdf}

\AppendGraphicsExtensions{.tif}

\begin{document}
\title{HOLMES: to Detect Adversarial Examples with Multiple Detectors} 
\author{Jing Wen}
\orcid{0003-4721-5327}
\authornote{This work was finished at the University of Hong Kong in 2020 and was also included in the author's dissertation\cite{wen2022defending}.}
\email{jwen@cs.hku.hk}
\affiliation{%
    \institution{The University of Hong Kong}
    \city{Dept. of Computer Science}
}

\begin{abstract}

Deep neural networks (DNNs) can easily be cheated by some imperceptible but purposeful noise added to images, and erroneously classify them. 
Previous defensive work mostly focused on retraining the models or detecting the noise, but has either shown limited success rates or been attacked by new adversarial examples.

Instead of focusing on adversarial images or the interior of DNN models, we observed that adversarial examples generated by different algorithms can be identified based on the output of DNNs (logits). Logit can serve as an exterior feature to train detectors. Then, we propose HOLMES (Hierarchically Organized Light-weight Multiple dEtector System) to reinforce DNNs by detecting potential adversarial examples to minimize the threats they may bring in practical.  HOLMES is able to distinguish \textit{unseen} adversarial examples from multiple attacks with high accuracy and low false positive rates than single detector systems even in an adaptive model. 
To ensure the diversity and randomness of detectors in HOLMES, we use two methods: training dedicated detectors for each label and training detectors with top-k logits. Our effective and inexpensive strategies neither modify original DNN models nor require its internal parameters. HOLMES is not only compatible with all kinds of learning models (even only with external APIs), but also complementary to other defenses to achieve higher detection rates (may also fully protect the system against various adversarial examples).

\end{abstract}



 
\maketitle

\section{Introduction}

Deep neural networks (DNNs) have demonstrated exceptional performance on various challenging artificial intelligence problems such as speech recognition \cite{speech} and recognition of voice commands \cite{voice2, voice,wu1}, natural language processing \cite{nlp,zhangsrame}, malware classification \cite{malware,malware2,malware3,wu2} and image recognition \cite{image,wu5}. Specifically, some DNNs \cite{imagen,wu4} can recognize images with an accuracy comparable to human.

However, recent research \cite{cw,adv,eae,delving,practical,limitations,wen2021,wu3} showed that DNNs used in image recognition is vulnerable to \textit{adversarial examples} \cite{adv}, which are unknowingly misclassified by DNNs. Usually, attackers add a small amount of noise that human can not notice to the correctly classified image (\textit{benign example}), and generate an adversarial example with a target label. By feeding the adversarial example to a DNN classifier, attackers can unknowingly take control of the classifier to produce the target label they want rather than the correct classification of the original images. Although the DNN model remains intact, adversarial examples intensively compromise its integrity.

Adversarial examples are significant threats to safety-critical systems using DNNs, such as automatic driving systems \cite{car,car2,drone,wu6}. For instance, attackers can add some imperceptible noise on a stop sign so that human recognizes it as a stop sign. However, it is misclassified as a yield sign by an autonomous car. As a result, the self-driving vehicle may not stop at the stop sign.

Our work aims to reinforce DNN systems against adversarial examples by detecting potential adversarial examples. Accurately identifying an attempted attack may be as important as recovering correct labels.
Even the most reliable defense cannot eliminate the adversarial examples in DNNs so far, it is a piece of cake to find adversarial examples. For security applications such as malware detection, attackers only need to find one successful adversarial example that is classified as benign to launch a malicious attack. Once it succeeds, all the defenses will collapse. Moreover, if adversarial examples have been detected, users may get alert and take control of systems to avoid unexpected behavior (e.g., a driver notices the stop sign and brakes) or systems take fail-safe emergency actions (e.g., an autonomous drone returns to its base). In an online scenario where DNNs are used as a service through API calls by external users, the ability to spot adversarial examples allows service providers to identify the potential attackers or malicious clients and reject their requests. 
\subsection{Our work}

How to theoretically eliminate the existence of adversarial examples in the deep neural network remains unsolved.
Given that DNNs are in widespread use at present, it is more realistic that we focus on how to detect them and minimize the threats they bring in practical.

The features of adversarial examples are keys to distinguish them. By analyzing the generation of adversarial examples and exploring why DNNs make wrong predictions for adversarial examples, we observed several significant differences in confidence scores distribution (also called logits) between benign and adversarial examples. The logit of the adversarial example has smaller maximum and variance than benign examples. 
Driven by our observation, logit can serve as an feature to train some binary detectors in advance to distinguish adversarial examples from benign examples.

\begin{figure}[hbtp]
\includegraphics[width = \linewidth]{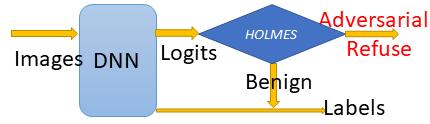}
\caption{HOLMES, as an external system, only requires logits from the DNN model.}
\label{f11}
\end{figure}

Besides, for some single detector systems, attackers can easily modify the adversarial examples slightly to bypass the detection\cite{bypass}. Training more non-differentiable detector can enhance the defence by adding extra randomness. To train various detectors to undertake the same detection task, we propose two training strategies: training dedicated detectors for each category and training detectors with top-k logits. These detectors constitute HOLMES, an external system to detect adversarial examples. The workflow is as simple as feeding the logit to HOLMES, and it reports whether it is adversarial, showing in Fig. \ref{f11}. Besides, HOLMES has the following features:

\begin{itemize}
\item \textbf{Accuracy.} HOLMES is highly accurate. Our evaluation exhibits it can detect more than 99\% adversarial examples with lower false adversarial rates. The AUCs are more than 0.97 even for unseen attacks.
\item \textbf{Transferablity.} Our evaluation shows it can be transferred to recognize adversarial examples from other  state-of-the-art attacks based on learning only two attacks . The accuracies for unseen adversarial examples are almost as same as the 2 attacks that were used in training.

\item \textbf{Compatibility.} HOLMES is compatible with various DNN models with different architectures since logit is the output of the last hidden layer, not the internal parameter.
\end{itemize}

\subsection{Contributions}
In summary, this paper makes the following contributions:
\begin{enumerate}
\item We characterize adversarial examples in terms of confidence scores distribution through a statistic method in Sec.\ref{char}. Their logits have relatively smaller maximum and lower variance than benign examples. As a result, different adversarial examples are easy to be classified by the same detector.

\item We propose a flexible and compatible system that can be deployed based on the intensity of defense in need. For training multiple detectors, we show two feasible methods in Sec.\ref{dvs}: training dedicated detectors for each label and training detectors with top-k logits.

\item We evaluate HOLMES on three benchmark datasets: MNIST, CIFAR-10, and ImageNet. We use the closed-world setting: training it with limited known attacks and testing it with various unknown adversarial examples generated from 8 different attack algorithms. This also shows the transferability of our detectors to some degree.
\end{enumerate}

Our experiment results in Section \ref{eva} demonstrate that HOLMES achieves more than 99\% detection rate for adversarial examples on three datasets with low false adversarial rates in a standard attack model. For an adaptive attack model where attackers know our defense exactly and seek means to bypass it, we attempted two adaptive methods. For bypassing only one detector, the attacks need to add 10 times more noise. Meanwhile, these adaptive examples can be detected by other detectors in HOLMES. For bypassing half the detectors, the adaptive attack fails to converge after the maximum iterations. The second way is increasing the confidence of adversarial examples.
Although the accuracy drops a little for high confidence adversarial examples. We propose another adaptive countermeasure to improve the detection rate back to 99\% again.

\section{Background and related work}

\subsection{DNNs for image recognition}
We introduce the following relevant sets, parameters, and definitions for deep neural networks used for image recognition:

\begin{itemize}
\item Input space $\mathbb{I}$: the set contains all the eligible input images, in which each pixel is an integer ranging from 0 to 255. In this paper, we normalize them into real numbers between -0.5 and 0.5.

\item Label space $\mathbb{L}$: the set includes mutually exclusive $k$ classes. For example, MNIST and CIFAR-10 are both class 10 examples, then $\mathbb{L} = \{0, 1, \dots, 9\}$.

\item Logit $\vec{y}$: is a $k$-dimensional vector from the last hidden layer. $\vec{y}_{i}$ denotes the $i$th value in $\vec{y}$, which is the confidence score of the $i$th classification.

\item Softmax $\vec{p}$: is also a $k$-dimension vector. We use $\vec{p}_{i}$ to denote the $i$th value in $\vec{p}$, which satisfies $ 0 \leq \vec{p}_{i} \leq 1, \sum_{i=1}^{k} \vec{p}_{i} = 1$. $\vec{p}_{i}$ represents the probability that $\vec{x}$ belongs to the $i$th class.

\end{itemize}

\begin{definition}
\label{logit}
The computational process from the input layer to the last hidden layer is a function: $h(\vec{x}) = \vec{y}$, which accepts $\vec{x} \in \mathbb{I}$ as input and generates the confidence score for each classification. 
\end{definition}

\begin{definition}
Softmax function is a monotonically increasing normalized function:
\begin{equation}
\vec{p}_{i} = \frac{e^{y_{i}}}{\sum_{j=1}^k e^{y_{j}}} \ \ \ \ for\ \  i = 1, 2, \dots, k
\end{equation}
\end{definition}

\begin{definition}
\label{dnn}
A deep neural network is a classifier: $l = f(\vec{x}) = argmax_{i}\vec{p}$, that chooses the $i$ with the largest probability to be the predicted label.
\end{definition}

\begin{definition}
The ground-truth classifier represents the label that human perceives: $l = g(\vec{x})$. Ideally, the perfect DNN classifier should predict a label that is close to human perception.
\end{definition}

Since softmax function is a monotonically increasing normalized function, the highest probability is equivalent to the highest confidence score. Therefore we can regard function $f$ in Def.\ref{dnn} as $l = f(\vec{x}) = argmax_{i}\vec{y}$. In this paper, we use logits $\vec{y}$ as the output of DNN and calculate the label by $l = argmax_{i}\vec{y}$. 

\subsection{Adversarial examples}
$\vec{x}' \in \mathbb{I}$ is an adversarial example originating from the benign example $\vec{x}$.

\begin{definition}
\label{advdef}
A successful targeted adversarial example $\vec{x}'$ for targeted label $t$ should satisfy the following constraints:
\end{definition}
\begin{enumerate}
\item $f(\vec{x}') = t, t \neq f(\vec{x})$, which makes DNN classifier generated the targeted label rather than the ground-truth label.
\item $g(\vec{x}') = f(\vec{x})$, this indicates that the adversarial example still looks like the benign example so that human is unable to perceive the adversarial example.
\end{enumerate}

\begin{definition}
A successful untargeted adversarial example $\vec{x}'$ should satisfy: $f(\vec{x}') \neq f(\vec{x})$ and $g(\vec{x}') = f(\vec{x})$.
\end{definition}

\subsection{Distance metrics}
For minimizing the noise added to the images, the distance between two image arrays is used to quantify the noise. There are three widely-used distance metrics to measure the distance between $\vec{x}$ and $\vec{x}'$:

\begin{enumerate}
\item $L_{0}$ distance reports on the total number of pixels that have been changed between $\vec{x}$ and $\vec{x}'$ no matter how much these values differ. It counts the number of $i$ such that $x_{i} \neq x_{i}'$.
\item $L_{2}$ measures the standard Euclidean distance between $\vec{x}$ and $\vec{x}'$: $\sqrt{\sum_{i}(\vec{x}_{i}-\vec{x}_{i}')^{2}}$. Unlike $L_{0}$ distance, $L_{2}$ distance can remain small even a large number of  pixels are changed.
\item $L_{\infty}$ measures the maximum change in any pixel. It only reports on $max\{|x_{1}-x_{1}'|, \cdots,|x_{m}-x_{m}'|\}$ even all the pixels in the image have been changed no more than that amount.
\end{enumerate}

No distance metric is optimal to measure human visual perception of two similar images. Of course, the smaller distance is, the less likely the adversarial example is to be noticed. In order to meet the second constraints in Def. \ref{advdef}, each attack just aims to minimize one of these three distance metrics. 

\subsection{Existing adversarial attacks}
\label{attacks}

Adversarial attacks transform images into adversarial examples by adding a small amount of noise in order to make it misclassified by DNN. Szegedy et al. \cite{adv} pointed out the existence of adversarial examples in DNNs for image recognition.
In general, targeted adversarial attacks were transformed into solving the following optimization problem theoretically.

\begin{equation}
\begin{array}{c}
minimize:\ D(\vec{x}, \vec{x}')\\
subject\ to:\ f(\vec{x}')=t, \ \vec{x}'\in \mathbb{I}
\end{array}
\label{e01}
\end{equation}
where $t$ is the targeted label and $D$ is a distance metric defining the distance between a pair of inputs.

\subsubsection{Fast Gradient Sign method (FGSM) \cite{eae}} 
Instead of searching for closest adversarial examples, FGSM is designed to generate adversarial examples as fast as possible. Intuitively, FGSM first calculates the gradient of loss function in order to determine in how to change (increase or decrease) intensity of each pixel to minimize the loss function. Then, all the pixels are changed simultaneously. Given a target label $t$ and input $\vec{x}$, it sets the Eq. \ref{eq3}, where $\epsilon$ should be set sufficiently small.
\begin{equation}
\vec{x}' = \vec{x} - \epsilon \cdot sign( \nabla loss_{f,t}( \vec{x}))
\label{eq3}
\end{equation}

\subsubsection{Jacobian-based Saliency Map Attack (JSMA) \cite{jsba}}
Papernot et al. proposed JSMA the under $L_{0}$ distance metric. JSMA is a greedy algorithm that makes use of saliency map to pick pixels to change. JSMA iteratively chooses pixels to modify until generating the targeted adversarial examples or reaching the maximum of iterations. They make use of the gradients of $h(\vec{x})_{t}$ to compute the saliency map, in which larger value indicates this pixel is more related to the image classified as the target label. According to the saliency map, they iteratively choose the pixels and alter it so that the image is more likely to be classified with label $t$. Besides, the author proposed another version of JSMA that makes use of $\nabla softmax(h(\vec{x}))_{t}$ in their defensive distillation \cite{dstl}.

\subsubsection{CW attacks}
Carlini et al. \cite{cw} first proposed a set of attacks under the three distance metrics. Their attack achieves very high success rates.

\textit{$L_{2}$ norm:}
They first make use of the $tanh$ function ($ -1 \leq tanh(w_{i}) \leq 1$) to automatically set $\vec{x}' = \frac{1}{2}(tanh(w)+1)$ in range $[0,1]^{m}$.
Given input $\vec{x}$ and target label $t$: 
\begin{equation}
\label{cwa}
\begin{array}{c}
minimize: ||\vec{x} - \vec{x}'||^{2} + \ c \cdot F( \vec{x}')\\
F( \vec{x}') = max(max\{h(\vec{x}')_{i}:i \neq t\} - h(\vec{x}')_{t}, - \kappa)
\end{array}
\end{equation}
Noted that $F( \vec{x}')$ is not the classifier function $f( \vec{x}')$, it is designed in such a way that $F( \vec{x}') \leq 0$ if and only if $\vec{x}'$ is wrongly classified with label $t$, which indicates the attack succeeds. Moreover,  $\kappa$ controls the confidence that $\vec{x}'$ is classified with label $t$. The larger $\kappa$ is, the higher confidence the adversarial example has, and the more noises are added meanwhile. $\kappa$ is normally set to 0.

\textit{$L_{0}$ norm:} 
Their $L_{0}$ attack is essentially an iterative algorithm similar to JSMA. In each iteration, they make use of their $L_{2}$ attack as well as the gradient of $F(\vec{x}')$ to determine the less important pixels for the classification $t$ then cut them until obtaining a small subset of pixels that can be changed to produce an adversarial example classified as label $t$. 

\textit{$L_{\infty}$ norm:}
They used $\vec{x}'=\vec{x}+\delta$, where $\delta$ is noises to be minimized. Then transform Eq. \ref{e01} into:
\begin{equation}
minimize: c\cdot F(\vec{x} + \delta) + \sum _{i}max\{(\delta _{i} - \tau),0\}
\end{equation}
To solve this optimization problem, it first begins with a small constant $c$ and doubles it in each iteration until generating a adversarial example or reaching the iteration maximum. Given each constant $c$, it reduces $\tau$ from 1 by 0.1 in every step until there exists any $\delta _{i} > \tau$.

There are other attacks such as L-BFGS \cite{adv}, Deepfool \cite{df}, Iterative Gradient Sign Method (IGSM) \cite{ifgs} and so on. We suggest readers referring to the original papers to get their ideas for attacks.

\section{Related defensive work}

The defensive work can be grouped into two broad categories: defensive methods \cite{adv,albad,magnet,dstl} that aim to enhance the network so that it correctly classifies adversarial examples with the ground-truth label, and detection methods \cite{dasfa,twins,statistical,cvl2}, which mainly focus on distinguishing the adversarial examples from benign examples. We introduce some of these methods below.

\subsection{Defensive methods}

Goodfellow et al. \cite{adv} proposed to re-train the DNN using the dataset including adversarial examples with the right label. Madry et al. \cite{albad} reported that, with adversarial learning, the classification accuracy drops to 87.3\% from 95.2\% on CIFAR-10.
Papernot et al. \cite{dstl} proposed the defensive distillation to train another DNN with the soft labels leveraging distillation training techniques \cite{distilling}. 
Carlini et al. \cite{cw} found that distillation cannot remove adversarial examples. The attacks still achieve 100\% success rates in their datasets.
Meng and Chen \cite{magnet} proposed MagNet composing of detector networks and reformer networks. Detectors learn the difference between benign and adversarial examples by approximating the manifold of benign examples. While reformers move the adversarial examples towards the manifold of benign examples to correctly classify them.
Cao and Gong \cite{region} proposed Region-based Classification (RC), which samples some nearby points in the hypercube centered at the testing example and re-uses DNN to predict the label for each sample point. Afterward, it ensembles all the labels and takes a majority vote among them as the final label. It seems the classifier makes a prediction for the region instead of the single point.

\subsection{Detection methods}

MagNet \cite{magnet} used autoencoders \cite{coder} to transform the image and comparing it with the original image. If the difference is larger than precomputed thresholds, MagNet decided it to be adversarial. Feature squeezing \cite{feature} shared the same idea with MagNet but used feature squeezing and smoothing method in computer vision to transform the original image and also set thresholds to decide if the image is adversarial. Wen et al. \cite{dcn} proposed a Detector-Corrector network (DCN). They built a single detector to distinguish the adversarial examples and a corrector to recover right labels. Logit pairing\cite{alp} encourages the logits from two images to be similar to each other.
LID\cite{lid} assesses the space-filling capability of the region surrounding a reference example. Using the distance distribution of the example to its neighbors measures adversarial subspace and distinguishes adversarial examples.

\subsection{Limitations of traditional methods}
Although previous work could detect adversarial examples to some extent.
They concentrated so much on the difference between benign images and adversarial images. Therefore, these methods have two key limitations:

\begin{enumerate}
\item Most methods require the whole or most of training dataset to train detectors in order to set proper thresholds which can pass most benign examples. Therefore, it is difficult to apply MagNet on ImageNet, which contains enormous training data. In addition, the training process for the whole dataset is time-consuming.
\item They are all vulnerable to adaptive adversarial attack since they use precomputed thresholds. Attackers are able to modify their loss function according to the defense and construct new adversarial examples to bypass the detection. Carlini et al. \cite{bypass} demonstrated that some previous detection methods failed to detect their new adversarial examples.
\end{enumerate}

\section{HOLMES}
\label{ho}
\subsection{Features of adversarial examples}
\label{char}

In Section \ref{attacks}, attacking algorithms make use of the gradient of function either $h$ or $f$ to determine the direction where the label moves towards the targeted label. Specifically, recalling the function $F( \vec{x}')$ in Eq. \ref{cwa}, it indicates the attack succeeds if and only if the confidence score of the targeted label $t$ becomes the maximum.
Therefore, adversarial algorithms mainly aim to increase the confidence score of the targeted adversarial label to the largest and reduce the confidence score of original label meanwhile. This process also perturbs the confidence scores of other classifications. But the perturbation is negligible since the final label is unilaterally determined by the maximum value.

\begin{figure}[hbtp]
\centering
\includegraphics[width = \linewidth]{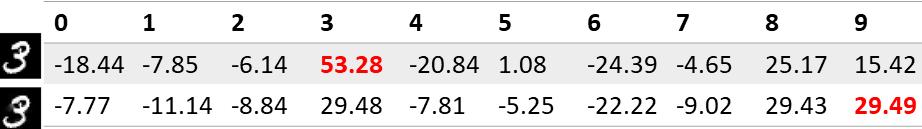}
\includegraphics[width = \linewidth]{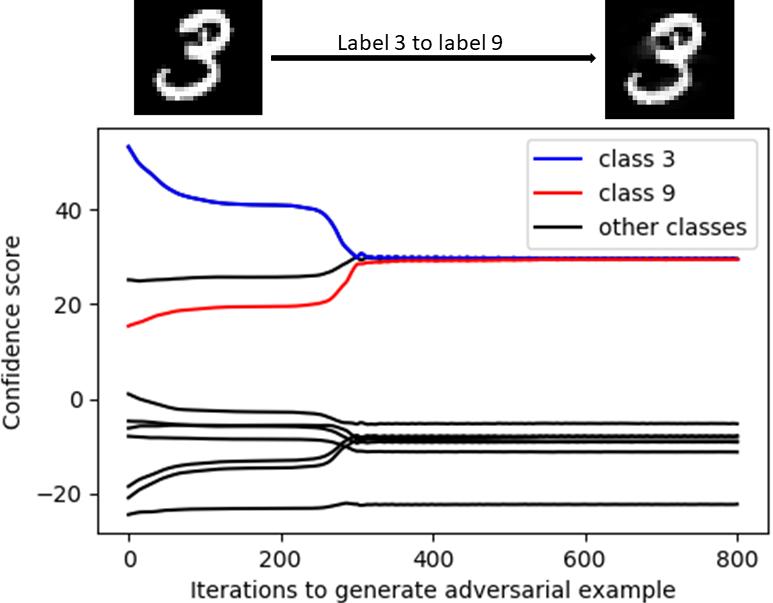}
\caption{How confidence scores change during the generation of adversarial 9 from benign 3. We use default CW-$L_{2}$ attack. The class 3 (blue line) is dropping during the procedure while the class 9 (red line) is increasing. The 8 black lines are other classes.}
\label{f2}
\end{figure}

\begin{figure*}[hbtp]
\includegraphics[width = \linewidth]{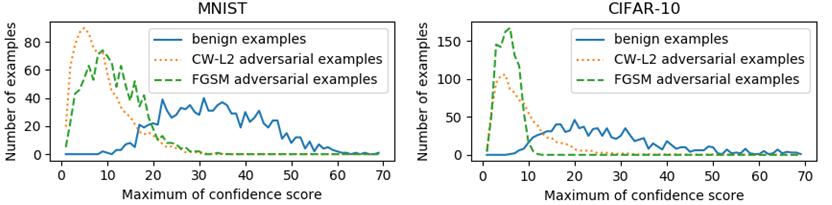}
\caption{Differences in maximums of confidence score for benign examples and adversarial examples. Each curve is fitted with histogram bins each representing the range of 1. Adversarial examples seem to have lower maximum confidence score than benign examples.}
\label{fstat1}
\includegraphics[width = \linewidth]{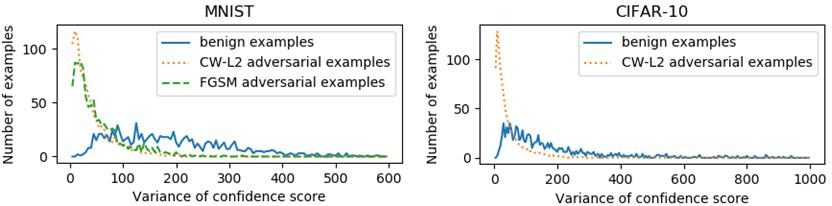}
\caption{Differences in variances of confidence score for benign examples and adversarial examples. Each curve is fitted with histogram bins each representing the range of 5. The confidence score distributions of adversarial examples are more concentrated since they have lower variances. }
\label{fstat2}
\end{figure*}

Driven by the algorithm, we record the confidence score during the generation process of an adversarial example and draw them in Fig. \ref{f2}. Also, we list the exact confidence scores before and after the attack. At the beginning, the benign example 3 has the largest confidence score $\vec{y}_3 = 53.28$ of label 3 while $\vec{y}_9 = 15.42$. With the increase in the number of attack iterations, $\vec{y}_3$ starts to decrease while $\vec{y}_9$ intends to increase until it becomes the maximum in $\vec{y}$.

Based on the changes in $\vec{y}$ , we initially consider that the logit of an adversarial example may have a lower maximum value and a more concentrated distribution(lower variance) than a benign example. Fig. \ref{fstat1} and Fig. \ref{fstat2} confirm our intuitions visually by showing the distribution of maximum and variance of logits for benign date, CW-$L_{2}$ (targeted) attacks and FGSM (untargeted) attacks. We randomly sample 1,000 benign data in MNIST and CIFAR-10 respectively and generate 1,000 successful adversarial examples for each attack in the default setting.

Fig. \ref{fstat1} shows the histogram of the maximum value of each logit for benign examples and two different adversarial examples (targeted and untargeted) on MNIST and CIFAR-10. The obvious peaks of two kinds of adversarial examples indicate most of the maximum confidence scores of adversarial examples are lower than 10 while those of benign examples range from 10 to 70. This statistic result is consistent with Fig. \ref{f2}: adversarial examples seem to have lower maximum confidence scores than benign examples.

In addition, we also observe that the confidence score distributions of adversarial examples are more concentrated. So we calculate the variances of each logit and show the histogram of them in Fig. \ref{fstat2}. For MNIST, most of the variances of adversarial examples are distributed within 100 and concentrated around 10 but the variances of benign examples are relatively evenly distributed below 400. We observe the same distribution for CIFAR-10 dataset. Actually, all the 1,000 data are lower than 10 so that the curve of FGSM adversarial examples begins at a very high point and drops to zero quickly. In that scale, it is hard to observe the other two curves so this curve is omitted in the histogram. 

Based on these statistic results, we suggest the following features of adversarial examples, which also motivate the design of the detector.

\begin{itemize}
\item \textit{Diffidence}: DNN classifies a wrong classification but with a lower confidence scores than usual. This is because the maximum in confidence scores of adversarial examples might be lower than benign examples' since the attack algorithm intends to decrease $\vec{y}_b$ (of the benign label) and increase $\vec{y}_t$ (of the targeted label) in the same time.
\item \textit{Concentration}: the confidence score distribution of adversarial examples might be more concentrated than benign examples' due to the decrease in maximum.
\end{itemize}

\subsection{Light-weight detectors}

We adopt the primitive idea of the detector in DCN \cite{dcn} to build a binary classifier and improve it to diverse detectors. 
Moreover, we firstly show the statistic results in Section \ref{char} to support this detection strategy since DCN just listed the experimental data but didn't provide any theoretical or statistic proof. Intuitively, we construct a three-layer (fully connected layer with ReLu) neural network classifier: $d(\vec{y}) = \{0,1\}$, to learn these differences in logits and decide whether the input is benign or adversarial.

Moreover, we find many advantages of using logit detectors. First of all, the traditional mathematical modeling often needs to quantify the whole dataset. While building classifiers only requires several samples(a small portion of a dataset) for training, which also improve the efficiency by the way. A well-trained classifier can perform consistently both on training data and unseen data from new attacks.
Secondly, building extra neural networks classifiers makes the whole system more robust to adaptive adversarial examples. Since current adversarial attacks only attack single network. With the protection of extra classifier $d$, a successful adversarial example should satisfy the extra constraint: $d(\vec{y}') = 0$ (0 represents benign examples) on the basis of $f(\vec{x}') = t, t \neq f(\vec{x})$. The existence of detectors increases the complexity of adaptive adversarial attacks to solve the optimization problems with the extra constraint.

\begin{figure}[hbtp]
\includegraphics[width = \linewidth]{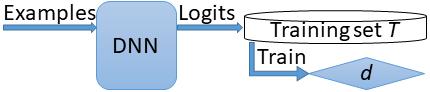}
\caption{Training phase of the detector. The detector is trained with the logits of images and labeled with 0 (benign) or 1 (adversarial).}
\label{f3}
\end{figure}

To train detectors, we randomly select benign examples from a dataset that DNN correctly classified and generate adversarial examples for them and represent the set as $\mathbb{E}$. Different from adversarial training, that is trained with examples $\mathbb{E}$, the detector is trained with the logits of them: $\mathbb{Y}=h(\mathbb{E})$, which is the output of DNN. We label the logits of benign examples with 0 while 1 represents adversarial. That's we get training dataset $\mathbb{T}$ for the detector. Fig. \ref{f3} shows the training phase of the detector.

\subsection{Diversification}
\label{dvs}

Intuitively, the more detectors are deployed in the system, the harder it is for attackers to bypasses all detectors. 
Consequently, We propose two improvements for a single detector to diverse detectors.
Diversification of a detector neither modifies the architecture of detector nor requires the extra data. We only make several strategic adjustments to the training dataset $\mathbb{T}$ in order to obtain different classifiers. As the diversification of detectors grows in HOLMES, it becomes harder for attackers to generate adaptive adversarial examples and provide more stable performance.

\subsubsection{Train dedicated detectors for each classification}

\begin{figure}[hbtp]
\includegraphics[width = \linewidth]{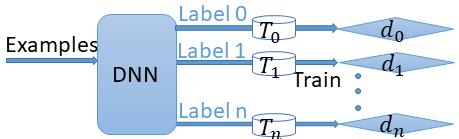}
\caption{Training phase of dedicated detectors.}
\label{f4}
\end{figure}

For some complex classification problems like CIFAR-100 with 100 classes, there is a brilliant strategy that first classifies the image into 20 superclasses. The idea is that classes within the same superclass are similar. Then it builds another network to classify the image within the same superclass.
Motivated by this strategy, we can build multiple detectors for each class. Since the DNN predicts the label for each example, the training dataset $\mathbb{T}$ is divided into $n$ classes spontaneously according to their label predicted by DNN: $T_{0}, T_{1}, \dots$. We use $d_{i}$ to represent the dedicated detector training with $T_{i}$ and will be especially used to distinguish data labeled with $i$ by DNN. Hence the single detector is replaced by $n$ dedicated detectors for each classification. Fig. \ref{f4} shows the training phase of multiple detectors. 

\subsubsection{Train detectors with top-k confidence scores}

Some networks for solving large-scale classification task (more than 1,000 classes) are measured by top-k accuracy, that is the probability that the correct label is one of the $k$ most likely labels reported by the network (e.g. Inception V3 training with ImageNet reported the top-5 accuracy). 
This indicates that the top-k confidence scores are more significant for classification. Meanwhile, in Fig. \ref{f2}, the fluctuations of $\vec{y}_3$ and $\vec{y}_9$, which are in top-3, are more noticeable than others. These ideas motivate us to train detectors learning the divergences of top-k confidence scores, which are more likely to occur between benign examples and adversarial examples.

Based on the training dataset $\mathbb{T}$, we use the top-k confidence scores to train detector. We use
$\mathbb{T}^{k}$ and $d^{k}$ to denote the new dataset and diverse detector.
In other words, we reduce the dimension for each item in $\mathbb{T}$ by discarding the smaller value it to form new train dataset.
In practice, this diversification increases the accuracy as well as reduce the training time.

\subsection{Hierarchical Organization}
\label{org}
In HOLMES, different detectors may disagree on the same adversarial example (e.g. attackers fool some detectors that predict the adversarial example as benign while rest of detectors think it's adversarial). We propose three policies for detectors in HOLMES to achieve consensus.

\begin{enumerate}
\item \textit{Any adversarial:} as long as there is any detector that reports one example to be adversarial, it's considered to be adversarial. 
\item \textit{Major adversarial:} HOLMES chooses the major decision among all detectors. If more than half of the detectors report examples is adversarial, HOLMES considers it to be adversarial. Otherwise, it is benign. 

\item \textit{All adversarial:} an example will be considered as benign unless all the detectors report it's adversarial.
\end{enumerate}

In practical, we can choose the most appropriate strategy to deploy HOLMES based on the need for sensitivity and specificity.

\section{Experimental setup}
Before the evaluation, we introduce the three benchmark datasets we use, describe the corresponding models, and set up different attack algorithms.
We use a PC equipped with an i5 3.40GHz CPU, 16GB RAM, and a GeForce GTX 1080 GPU to conduct the following experiments.

\subsection{Datasets and DNN models}

MNIST \cite{mnist} is a dataset of handwritten digits (0 to 9) formatted as $28\times28$ in gray scale. CIAFR-10 \cite{cifar} collected 10 kinds of tiny color images formatted as $32\times32\times3$. 
Along with considering the small size MNIST and CIFAR-10, we also consider the ImageNet dataset, which contains 1.2 million training images of different size.
The ImageNet \cite{imagenet} dataset is provided by ImageNet Large Scale Visual Recognition Challenge 2012 for a 1,000-classification task. We use the pre-trained Inception V3 \cite{inception} for ImageNet, which takes images formatted as $299\times299\times3$ and predicts 1,008 classes (those from 1 to 1,000 are valid classifications, the rest is for compatibility with old versions). 
Table \ref{t3} lists the training accuracy, testing accuracy for MNIST and CIFAR-10 and top-5 accuracy (the probability that the correct class is one of the five most likely classes reported by the model) for ImageNet. Our models all achieve comparable accuracy \cite{result} with the state-of-the-art network architectures. Noted that the CIFAR-10 model significantly overfitted the training data even with adding dropout layer with a rate of 0.5, we did not modify the DNN model since CW-attack did not optimize it either.

\begin{table}[htbp]
\caption{Accuracy of DNN models on three datasets}
\centering
\begin{tabular}{l|c|c|c}
\hline
Accuracy       &MNIST    &CIFAR-10 &ImageNet\\
\hline
Training       & 99.98\% & 99.74\% & -      \\
Testing (Top-1)& 99.42\% & 77.96\% &78.8\%  \\
Testing (Top-5)&  -      &  -      &94.4\%  \\
\hline
\end{tabular}
\label{t3}
\end{table}

\subsection{Adversarial examples}

We test HOLMES against open-source CW attacks from authors \cite{cw}, FGSM, and JSMA provided by CleverHans \cite{cleverhans}. 

\subsubsection*{CW attack}
We test a set of targeted CW attacks with default parameters since CW attacks can generate adversarial examples under three distance metrics on three benchmark datasets. In addition, we transform the targeted attacks into corresponding untargeted attacks.
For CW-$L_{2}$ attack, we randomly choose 500 benign examples that DNN correctly classifies from each dataset, and generate adversarial examples targeting at the rest 9 labels for MNIST and CIFAR-10. For ImageNet, we target at random 9 labels between $[1,1000]$ except the right one using CW-$L_{2}$. As for CW-$L_{0}$ and $L_{\infty}$ we only use 100 benign examples since these two algorithms are too inefficient to run plenty of samples. Besides, it is expensive to generate adversarial examples for ImageNet under $L_{0}$ and $L_{\infty}$ since it took more than 10 minutes for each one. 

\subsubsection*{FGSM}
Since we use the implementation of untargeted FGSM, we first generate 1,000 adversarial examples for 1,000 random benign examples on two datasets. But we find the success rate is too low on MNIST then we generate another 2,000 examples and eliminate the unsuccessful examples. 

\subsubsection*{JSMA}
is a targeted attack. We randomly choose 100 benign examples to generate 900 adversarial examples on MNIST and CIFAR-10 and eliminate the unsuccessful examples.

\subsection{HOLMES setup}

For training detectors in HOLMES, we only use $CW-L_{2}$ and $FGSM$. First, we randomly select 1,000 benign examples from MNIST that DNN correctly classified and generate adversarial examples for them. Specifically, for each benign example, we use CW-$L_{2}$ to generate 9 targeted adversarial examples. Besides, we use FGSM to generate untargeted examples and eliminate the unsuccessful examples. Finally, we obtain datasets composed around 10,000 examples in total. For CIFAR-10 and ImageNet, we reduce the training dataset size to half since the single image needs more time to process, we want the training procedure to be as fast as possible. So the training set contains about 5,000 examples respectively. We implemented HOLMES in Python 3.6.2 using Keras \cite{keras} and Tensorflow \cite{tensorflow} as backend. 

\subsubsection*{Diverse detectors}

For MNIST and CIFAR-10, we first divide the dataset into 10 subsets and then use each of them to train a set of dedicated detectors: $d_{0}, d_{1},\dots, d_{9}$. Since for every example, only one dedicated detector with the same classification will work. We use $d_{*}$ to represent all the dedicated detectors and regard them as one detector in the following. Besides, we train two top-k detectors: $d^9$ and $d^{8}$. So we build 3 detectors in all (12 detectors in reality).
Note that we only aim to train a basic classifier to show how well HOLMES distinguish adversarial examples. Based on our preliminary experiment, three detectors are adequate. How the number of detectors impact the performance will be left as future work.

It is cumbersome to apply $d_{*}$ diversification to the large-scale classification tasks like ImageNet as it is inefficient to build 1008 dedicated detectors. In practice, we can build 10 detectors and each of them takes charge of 100 classifications. But in our evaluation, we only build three top-k (top-20, top-40, and top-60) detectors for ImageNet since it is more efficient and practical.

\subsection{Threat models}
\begin{itemize}
\item \textit{Standard attack model:} we assume that attackers always knows everything about DNN classifier $f$, such as its architecture, parameters, training dataset, training procedure. They treat attacked DNN classifier as a white box. Since the defense doesn't change anything about DNN classifier, attackers do not notice the existence of detector $d$.
\item \textit{Adaptive attack model:} based on the standard attack model, attackers knows everything about the detector $d$ such as its architecture, parameters and so on. In this scenario, a simple adaptive strategy is to generate adversarial examples with higher confidence. On the other hand, attackers can modify their attack algorithms to bypass our detectors. 
\end{itemize}

\section{Evaluation}
\label{eva}
\subsection{Against standard attacks}

\subsubsection{False adversarial rate} measures the proportion of benign examples that are incorrectly identified as adversarial. If a detector detects most adversarial examples at the expense of plenty of benign examples as adversarial, it is entirely useless. Therefore, false adversarial rate is an important indicator. For MNIST and CIFAR-10, we test the whole training and validation data excluding the data used in training detectors. While for ImageNet we test 5,000 images picked from 10 classifications randomly. We never test HOLMES with the data used in the training phase to guarantee the authenticity of accuracy.

\begin{table}[htbp]
\centering
\caption{False adversarial rates for benign examples}
\begin{tabular}{|l|c|c|c|c}
\hline
         &Any      &Major    &All    \\
\hline
MNIST    & 3.17\% & 1.64\% & 0.78\% \\
CIFAR-10 & 15.95\% & 9.19\% & 4.96\% \\
ImageNet &  26.6\% &  7.4\% & 2.4\%  \\
\hline
\end{tabular}
\label{t4}
\end{table}

Table \ref{t4} lists the false adversarial rates of HOLMES. Noted that the detectors were only trained with around 2\% benign examples but still correctly classify most other unseen benign examples. As described in Section \ref{org}, these three policies hold different accuracies on benign examples. The Any policy performs worst and the Major policy and All policy both are more accurate. Actually, these three policies are different tradeoffs between accuracies for benign and adversarial examples. We will show the true adversarial rates and discuss these policies.

\begin{table}[htbp]
\centering
\caption{True adversarial rates and AUC ROC for different attacks (HOLMES is trained by CW-$L_{2}$ and FGSM)}
\begin{tabular}{|l|c|c|c|c|c|c|}
\hline
\multicolumn{3}{|c|}{}&Any&Major&All&AUC ROC\\
\hline
\multirow{8}{*}{\rotatebox{90}{MNIST}} & \multirow{4}{*}{\rotatebox{90}{Targeted}}
 &CW-    $L_{0}$ & 100\%   & 100\%    & 99.88\%& 0.9960  \\
&&CW-    $L_{2}$ & 100\%   &  99.97\% & 99.93\%& 0.9960  \\
&&CW-$L_{\infty}$& 100\%   & 99.89\%  & 98.56\%& 0.9959\\
&&JSMA           & 100\%   &  100\%   & 99.81\%& 0.9960 \\
\cline{2-7}
& \multirow{4}{*}{\rotatebox{90}{Untargeted}}
 &CW-     $L_{0}$&  100\% & 100\% & 100\% & 0.9961 \\
&&CW-     $L_{2}$&  100\% & 100\% & 100\% & 0.9961 \\
&&CW-$L_{\infty}$&  100\% & 100\% & 100\% & 0.9961\\
&&FGSM           &  100\% & 100\% & 99.74\%& 0.9960\\
\hline
\multirow{8}{*}{\rotatebox{90}{CIFAR-10}}& \multirow{4}{*}{\rotatebox{90}{Targeted}}
 &CW-     $L_{0}$& 100\% & 100\%   & 99.56\% & 0.9749 \\
&&CW-     $L_{2}$& 100\% & 99.95\% & 98.91\% & 0.9746 \\
&&CW-$L_{\infty}$& 100\% & 99.88\% & 98.56\% & 0.9744 \\
&&JSMA           &  100\%  & 100\%   & 99.87 & 0.9751\\
\cline{2-7}
& \multirow{4}{*}{\rotatebox{90}{Untargeted}}
 &CW-     $L_{0}$& 100\%   & 99\%  & 96\%    & 0.9728   \\
&&CW-     $L_{2}$& 100\%   & 99.8\%& 93.8\%  & 0.9722    \\
&&CW-$L_{\infty}$& 100\%   & 99\%  & 93\%    & 0.9714   \\
&&FGSM           &  100\%  & 100\% & 99.76\% & 0.9750   \\
\hline
\multirow{3}{*}{\rotatebox{90}{\tiny{ImageNet}}}
& \multicolumn{2}{c|}{\multirow{2}{*}{T-CW-$L_{2}$}}& \multirow{2}{*}{99.91\%} &\multirow{2}{*}{98.02\%} & \multirow{2}{*}{94.82\%} & \multirow{2}{*}{0.9832}\\
& \multicolumn{2}{c|}{}&&&&\\
& \multicolumn{2}{c|}{U-CW-$L_{2}$} & 100\%   &97.5\%  & 91.5\%  & 0.9818\\

\hline
\end{tabular}
\label{t5}
\end{table}

\subsubsection{True adversarial rate} measures the percentage of adversarial examples that are correctly identified as adversarial. 
Table \ref{t5} lists the true adversarial rates and AUC ROC to show the outstanding performance of HOLMES for detecting successful adversarial examples of different attacks. Each AUC represents the detection accuracy for both benign examples and adversarial examples from each attack as we have three policies, which are equivalent to different thresholds.
For each data set, even for different attacks, each AUC score is very close. It is worth mentioning that
the training set for HOLMES only contains adversarial examples generated by targeted CW-$L_{2}$ and untargeted FGSM. Although, HOLMES never know the examples generated from CW-$L_{0}$, CW-$L_{\infty}$ and JSMA.
It still can detect adversarial examples from unseen attacks with almost the same high accuracy. Therefore, HOLMES can be transferred to detector multiple adversarial examples.

For different policies, first of all, HOLMES with Any policy achieves almost 100 \% detection rate for adversarial examples on three datasets. Once a detector reports an example to be adversarial, HOLMES considers it to be adversarial. In the meanwhile, this policy sacrifices fasle adversarial rates for benign examples. This strategy is extremely strict and it is very suitable for systems that are required not to miss any adversarial examples. Secondly, the Major policy achieves more balanced performance. With lower false adversarial rate, the Major can detector almose the 100\% adversairal examples.
Thirdly, the All policy is a little worse than other policies but significantly decreases false adversarial rates listed in Table \ref{t4}. The All policy is designed for systems sensitive to benign examples. The results show that these three policies are different tradeoffs between false adversarial rate and true adversarial rate. Which policy to be used depends on the defensive intensity that one system needs. In general, the Major policy has the most stable and balanced performance. In the next discussion, we will mainly focus on Major policy.

\subsection{Comparison with other detection methods}

HOLMES is compared with the single detector in DCN and Feature squeezing. Table \ref{t77} lists some properties of these three methods. First of all, the single detector in DCN did not handle the ImageNet dataset on Inception V3 model since it contains 1008 classifications. It is unrealistic to get enough adversarial samples for training. In our improvement, we only use the top-k scores in each logit so we compress the training dataset and provide detectors with adequate knowledge. So, HOLMES is flexible to handle small or large problems. Secondly, we use CW-$L_{2}$ and FGSM to train and test with the new adversarial examples generated from these two and other new attacks. Actually, HOLMES performs stably against known and unknown attacks. This cross-validation proves that HOLMES is able to detect novel and unknown attacks. DCN only trained and tested with CW attacks and Feature squeezing had to run all the attacks samples to set proper squeezers together to achieve their best detection performance.
Thirdly, the training procedure of HOLMES is simple. We only need a small portion of benign data and corresponding adversarial examples to train while FS has to run the whole dataset and all attacks to set proper detection parameters.
Lastly, we test HOLMES with an adaptive attack model and show it is robust enough against adaptive attacks in the next section. 

\begin{table}[htbp]
\centering
\caption{Properties of DCN, Feature Squeezing (FS), and HOLMES}
\begin{tabular}{l|l|c|c|c|}
\hline
\multicolumn{2}{}{} & DCN &FS & HOLMES\\
\hline
\multirow{4}{*}{\rotatebox{90}{Properties}}&For larger scale problem& $\times$ & ${\surd}$&${\surd}$ \\
&Cross validation  &$\times$  &$\times$  &${\surd}$\\
&Set-up simplicity                 &${\surd}$ &$\times$  &${\surd}$\\
&Adversarial adaption              &$\times$  &${\surd}$ &${\surd}$\\
\hline
\multirow{3}{*}{\rotatebox{90}{AUC ROC}}&MNIST&0.9965&0.9974&0.9960\\
&CIFAR-10&0.9687&0.9524&0.9738\\
&ImageNet&-&0.9424&0.9825\\
\hline
\end{tabular}
\label{t77}
\end{table}

We also show calculate the AUC ROC of each detection method and listed in Table \ref{t77}. For MNIST, all three approaches are over 0.99. For larger dataset such as CIFAR-10 and ImageNet, the gap in accuracy appears. 
HOLMES is around 0.98 while other methods are lower.  Therefore, HOLMES performs stably with high accuracy over different datasets and DNN models than these traditional methods.

\subsection{Against adaptive bypassing attacks}

So far we demonstrated the extraordinary accuracy on detecting adversarial examples under a standard attack model. We now consider a more realistic scenario where the powerful attackers also know the existence of these effective detectors and immediately seek means to evade the detection. 
Since some adversarial attack such as CW-$L_{2}$ can achieve 100\% success rate for kinds of networks, there is no doubt that CW-$L_{2}$ can bypass each detector solely. Actually bypassing single detector is useless. A successful adaptive adversarial example must be misclassified by a DNN model as well as bypass detector networks simultaneously.
Thus we improve the CW-$L_{2}$ attack to bypass detectors.
Similar to the original adversarial attack, the adaptive adversarial attack is still a constrained optimization problem. For bypassing the known detectors, attackers need to take more constraints into consideration when generating the adversarial examples. For our detectors, attacks wants the output of each detector to be below 0.5, which means the judgement is benign. Mathematically, the adaptive adversarial attack can be formulated into:

\begin{equation}
\label{cwa}
\begin{array}{rl}
minimize: &||\vec{x} - \vec{x}'||^{2} + \ c \cdot F( \vec{x}')+\sum_i e_i \cdot G(\vec{x}')\\
F( \vec{x}') =& max\{h(\vec{x}')_{i}:i \neq t\} - h(\vec{x}')_{t}\\
G(\vec{x}') =& d_i[h(\vec{x}')]-0.5\\
subject\ to: & c,e_i > 0\ \ ,\ x'\in \mathbb{I}\\
\end{array}
\end{equation}

Based on the $CW-L_2$ attack, we construct a new adaptive adversarial attack for our detectors. For simplicity, we remove the confidence parameter $\kappa$, which is equivalent to setting $\kappa=0$. Since CW attack was reported that with the increase of confidence parameter, the success rate drops. Setting $\kappa=0$ also ensures that we can generate as many successful adversarial examples as possible. The objective function in Eq.\ref{cwa} includes not only the original neural network $f$, as well as the detector networks $d$. This attack algorithm becomes more complicated for each additional detector to be bypassed. Although there is more than one detector in HOLMES, we started the adaptive adversarial attacks with bypassing one detector. 

\begin{table}[htbp]
\centering
\caption{Success rates of adaptive adversarial attacks}
\begin{tabular}{|l|c|c|c|}
\hline
Dataset &Fool the network& Fool one detector& Fool both\\
\hline
MNIST    & 31.44\% & 89\% & 20.44\%  \\
CIFAR-10 & 65.39\% & 92.25\% & 57.65\% \\
\hline
\end{tabular}
\label{t631}
\end{table}

Table \ref{t631} lists the success rates of adaptive attacks for bypassing one detector. Even if the attackers know one of our detectors, the attack algorithm still performs poorly. In addition, our detectors not only reduce the success rate of attack algorithm but also force the attacker to add more noise, which makes these adversarial examples detectable by humans. For MNIST dataset, the success rate for fooling both the network and the detector is only 20.44\%. The average noise added in those successful adversarial examples is 11.06. In contrast, the original $CW-L_2$ achieved 100\% success rate with adding average 1.23 noise. For CIFAR-10, the adaptive attack achieves 57.65\% success rate with adding average 1.305 noise. Comparing with the average noise of 0.235 under a standard attack model, the adaptive attack has to add 5 times more noise to reach a limited success rate. 

\begin{table}
\caption{Detection accuracy of HOLMES for successful adaptive adversairal examples.}
\begin{tabular}{|l|c|c|c|}
\hline
  & Any&Major&All\\
\hline
MNIST    & 100\% & 97.88\% & 0\% \\
CIFAR-10 & 97.41\% & 87.53\% & 0\% \\
\hline
\end{tabular}
\label{t632}
\end{table}

Although we can design an adaptive attack algorithm for specific detectors, Table \ref{t632} shows these adaptive adversarial examples are still \textit{detectable} by the rest detectors in HOLMES, which is the one we used in the previous standard model and never know the new adaptive adversarial examples. Under All policy, HOLMES always fails since one detector has been bypassed. As for Any and Major policy, HOLMES is still effective and achieves high accuracy for adaptive attacks that can bypass one detector.

For bypassing more than one detector, we try different initial parameters and find the attack fails to converge after running the maximum iterations. Based on these poor-quality adaptive adversarial examples and high detection accuracy, we can safely conclude our HOLMES is robust enough against adaptive adversarial attacks which want to bypass our detectors. Moreover, the cost of bypassing one more detector greatly exceed the cost of training one more detector. The existence of HOLMES make the adaptive attack comes with the expense of lower success rate and more noise.

\subsection{Against adaptive high confidence attacks}

\label{61}

It seems adaptive attacks fail to bypass HOLMES. But we can try another direction to avoid the detection. CW-$L_{2}$ attack can easily increase the confidence parameter without changing the attack algorithm to generate stronger adversarial examples with higher confidence, which may differ from what detectors have learned. The higher confidence parameter is, the stronger classification confidence is, which makes it possible to bypass the detections since the detectors heavely rely on the confidence scores. However, 
the higher confidence examples are also flawed. With the increase of attack confidence parameter, the success rate actually decreases. Since the higher confidence brings larger noises on images, the adversarial example is more likely to be noticed by human and violate the ground-truth classifiers in Def. \ref{advdef}. In other words, the high confidence attack adds so many noises that adversarial images do not look like benign examples anymore like the adaptive bypassing method.

\begin{figure}[hbtp]
\includegraphics[width = \linewidth]{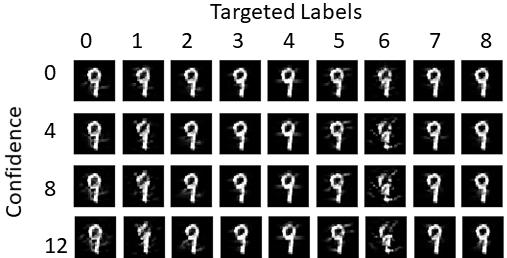}
\caption{The adversarial examples with different target labels and confidence parameters. They are generated from the same benign image labeled with 9.}
\label{f6}
\end{figure}

Fig. \ref{f6} shows the adversarial examples with different target labels and confidence parameters. They are all generated from the same benign image labeled with 9. Obviously, when attack confidence is 12, the adversarial examples targeted at 0, 1, 6 are easy to be noticed by human. So the confidence parameter should be moderate to avoid human perception. Actually, the average noise added on confidence 12 examples is 2.564, almost one times more than the noise added on confidence 0 examples.

In our evaluation, we generated 900 adaptive adversarial examples for each confidence parameter:0,2,...,12. We use the \textit{identical} detectors in the previous evaluation. Noted that our detectors are trained with the adversarial examples with confidence 0. They never see the higher confidence adversarial examples.

\begin{figure}[hbtp]
\includegraphics[width = \linewidth]{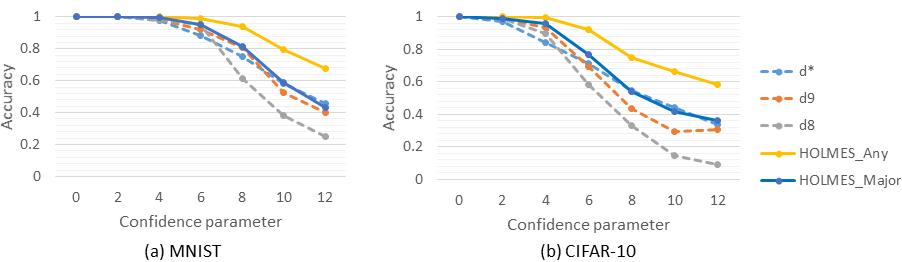}
\caption{How detection accuracy changes with the increase of confidence parameter of CW-$L_{2}$ attack.}
\label{f5}
\end{figure}

Fig. \ref{f5} shows the accuracy of each detector used in HOLMES and HOLMES with Any and Major policy. We omit the All policy as this policy provides the weakest detection and not suitable for high confidence adversarial examples. But we list the detailed accuracy of this three policy in Table \ref{t7} in the Appendix.

Firstly, with the increase of attack confidence parameter, the accuracy of single detectors and HOLMES both drops. Especially for CIFAR-10, one single detector drops from 100\% to 10\%. Since all these detectors are feed with confidence scores, increasing the attack confidence is indeed an effective way to bypass our detection. Secondly, HOLMES with Major policy performs better than all the single detectors. When attack confidence parameter is 12, HOLMES with Major policy drops to around 40\% while some single detector only detects 10\% adversarial examples. This is because HOLMES integrates information from diverse detectors so it provides more accurate judgment. Third, HOLMES with Any policy provides the most robust defense against high confidence adversarial examples. For MNIST, it remains more than 80\% accuracy until the parameter increases to 12 while keeps more than 90\% until attack confidence reaches 8 for CIFAR-10. In summary, HOLMES still performs better than single detectors in the adaptive attack model.

\subsection{Our countermeasure to adaptive high confidence attacks}
\label{62}

One straightforward strategy to combat adaptive attacks is to train HOLMES with high confidence adversarial examples. Since the detectors we use in the last section are trained with the adversarial examples with confidence 0. They never see the higher confidence adversarial examples. No wonder they fail to distinguish the adversarial examples with higher confidence. We use the same HOLMES parameter and settings but replacing the training examples with higher confidence 6 and 12 to see if this countermeasure works.
\begin{figure}[hbtp]
\includegraphics[width = \linewidth]{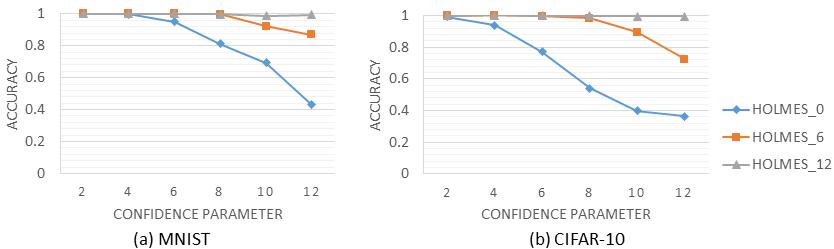}
\caption{How detection accuracy changes with the increase of confidence parameter of CW-$L_{2}$ attack for HOLMES with Major policy training on adversarial examples at different confidence level (0, 6, 12).}
\label{f7}
\end{figure}
Fig. \ref{f7} draws three curves and each shows the HOLMES with Major policy training with adversarial examples at different confidence level (0, 6, 12). As we stated in the previous section, the Major policy is the most stable policy both for adversarial examples and benign examples despite that the Any policy could provide higher accuracy for high confidence parameter. So we only show the Major policy curves in Fig. \ref{f7}. We list the detailed accuracy of three policy in Table \ref{t7} in the Appendix. 

The results in Fig. \ref{f7} meet our expectations. If we train HOLMES with higher confidence adversarial examples, the detection accuracy increases obviously. Secondly, HOLMES with higher confidence is compatible with the adversarial examples lower than the training confidence. For instance, although the HOLMES training with 6 confidence adversarial examples, it still can detect the adversarial examples at 2, 4 confidence level with 100\%. HOLMES training with 12 confidence adversarial examples is the same, it can detect almost all the adversarial example with confidence no more than 12.

Actually, this countermeasure is imperfect as we obverse the slight increase in false adversairal rate. How to train detectors to distinguish the high confidence adversarial examples from benign examples is an interesting future work.

\begin{table*}[hbtp]
\centering
\caption{The true adversarial rate of HOLMES training with adversarial examples at different confidence lever under three policy.}
\begin{tabular}{c|c|c|c|c|c|c|c|c|c|c}
\hline
\multirow{2}{*}{Dataset}&\multirow{2}{*}{Confidence}&\multicolumn{3}{c|}{HOLMES}&\multicolumn{3}{c|}{HOLMES 6 }&\multicolumn{3}{c}{HOLMES 12} \\
\cline{3-11}
&&Any&Major&All&Any&Major&All&Any&Major&All\\
\hline
\multirow{7}{*}{MNIST}
&2 &  100\%   & 100\%   &   100\% & 100\%   & 100\%   & 100\%   & 100\% & 100\%   & 99.89\% \\
&4 &  100\%   & 99.67\% & 95.78\% & 100\%   & 100\%   &  100\%  & 100\% & 100\%   & 100\%   \\
&6 &  99.11\% & 95.11\% & 81\%    & 100\%   & 100\%   & 99.22\% & 100\% & 100\%   & 100\%   \\
&8 &  94\%    & 81.22\% & 55.67\% & 100\%   & 99.78\% & 93.67\% & 100\% & 99.88\% & 99.33\% \\
&10& 86.31\%  & 69.42\% & 42.36\% & 99.6\%  & 92.27\% & 84.47\% & 100\% & 98.94\% & 98.31\% \\
&12& 67.33\%  & 43.11\% & 21.67\% & 96.22\% & 86.78\% & 64\%    & 100\% & 99.44\% & 96.11\% \\
\hline
\multirow{7}{*}{CIFAR-10}
&2 & 100\%   & 99.33\% & 94.44\% & 100\%   & 99.89\% & 99.56\% & 100\%   & 99.78\% & 99\%    \\
&4 & 99.33\% & 93.78\% & 78.56\% & 100\%   & 100\%   & 97.33\% & 100\%   & 100\%   & 99.56\% \\
&6 & 92.11\% & 77.11\% & 50\%    & 100\%   & 99.56\% & 96\%    & 100\%   & 99.78\% & 99.11\% \\
&8 & 74.89\% & 54.11\% & 24.78\% & 99.89\% & 98.33\% & 88.33\% & 100\%   & 99.67\% & 98.33\% \\
&10& 63.29\% & 39.67\% & 15.27\% & 97.27\% & 89.42\% & 66.16\% & 99.93\% & 99.53\% & 97.16\% \\
&12& 58.56\% & 36.33\% & 16.44\% & 87.56\% & 72.45\% & 46.56\% & 100\%   & 99.56\% & 94.11\% \\
\hline
\end{tabular}
\label{t7}
\end{table*}

\section{Conclusions}

HOLMES holds surprising effectiveness and accuracy. Compared to other previous defenses, it is such simple and inexpensive. How to theoretically eliminate the existence of adversarial examples in deep neural network remains unsolved, but our intuition focuses on how to detect them and minimize the threats they bring in practical.

In summary, we find logic can serve as a feature to train detectors to distinguish adversarial examples.
Then we propose HOLMES that assembles multiple detectors that can spot the unknown adversarial examples with high accuracy. Although we only implement three detectors, we show two general methods to train more detectors. Moreover, we propose two adaptive attacks aiming at HOLMES and provide an effective countermeasure against these adaptive adversarial examples. This technique doesn't need to explore the interior of networks and compatible with various models.
There are also several drawbacks of HOLMES: it requires several adversarial examples in the training phase, it can not recover the right label of adversarial examples. 

Future work includes exploring how the number of detectors will impact the performance, other diversification methods to train an effective detector and exploring other methods to train detectors. We encourage researchers who develop new adversarial attacks to evaluate their attacks against our HOLMES. To enable other researchers to test with our work in an easy manner, all of our implementations are available at: ( we temporarily hide the link for anonymity).

\section*{appendix}
Table \ref{t7} lists the detailed accuracies of HOLMES training with adversarial examples at different confidence level under three policy mentioned in Sec. \ref{61} and Sec. \ref{62}.


\bibliographystyle{ACM-Reference-Format}
\bibliography{sample}

\end{document}